\documentclass{article}
\usepackage{nodalida2021}

\usepackage[utf8]{inputenc}
\usepackage[english]{babel}
\usepackage{mathptmx}
\usepackage{caption}
\usepackage{subcaption}
\usepackage{graphicx}
\usepackage{soul}\setuldepth{article}
\usepackage{url}

\usepackage{listings}
\usepackage{todonotes}
\usepackage{verbatim}
\usepackage{multirow}

\usepackage{lipsum}

\graphicspath{{images/}{../images/}}
\usepackage{subfiles}

\usepackage{blindtext}
\usepackage{booktabs}
\usepackage{hyperref}

\hypersetup{
    colorlinks=true,
    citecolor=blue,
    filecolor=blue,      
    urlcolor=blue,
    linkcolor=blue
}

\title{EstBERT: A Pretrained Language-Specific BERT for Estonian}
\author{Hasan Tanvir \and Claudia Kittask \and Sandra Eiche \and Kairit Sirts \\
         Institute of Computer Science \\
         University of Tartu \\
         Tartu, Estonia \\
         \texttt{hasantanvir79@gmail.com}, \texttt{\{claudia.kittask,sandra.eiche,sirts\}@ut.ee} \\
         
}
\date{ }

\aclfinalcopy

\begin{document}

\maketitle

\begin{abstract}
   This paper presents EstBERT, a large pretrained transformer-based language-specific BERT model for Estonian. Recent work has evaluated multilingual BERT models on Estonian tasks and found them to outperform the baselines. Still, based on existing studies on other languages, a language-specific BERT model is expected to improve over the multilingual ones. We first describe the EstBERT pretraining process and then present the models' results based on the finetuned EstBERT for multiple NLP tasks, including POS and morphological tagging, dependency parsing, named entity recognition and text classification.
   The evaluation results show that the models based on EstBERT outperform multilingual BERT models on five tasks out of seven, providing further evidence towards a view that training language-specific BERT models are still useful, even when multilingual models are available.\footnote{The model is available via HuggingFace Transformers library: \url{https://huggingface.co/tartuNLP/EstBERT}}
\end{abstract}

\section{Introduction}

Pretrained language models, such as BERT \cite{devlin2019bert} or ELMo \cite{peters2018deep}, have become the essential building block for many NLP systems. These models are trained on large amounts of unannotated textual data, enabling them to capture the general regularities in the language and thus can be used as a basis for training the subsequent models for more specific NLP tasks.
Bootstrapping NLP systems with pretraining is particularly relevant and holds the greatest promise for improvements in the setting of limited resources, either when working with tasks of limited annotated training data or less-resourced languages like Estonian.

Since the first publication and release of the large pretrained language models on English, considerable effort has been made to develop support for other languages. In this regard, multilingual BERT models, simultaneously trained on the text of many different languages, have been published, several of which also include the Estonian language \cite{devlin2019bert,conneau2019unsupervised,sanh2019distilbert,conneau2019cross}. 
These multilingual models' performance was recently evaluated on several Estonian NLP tasks, including POS and morphological tagging, named entity recognition, and text classification \cite{kittask2020evaluating}.
The overall conclusions drawn from these experiments are in line with previously reported results on other languages, i.e., for many or even most tasks, multilingual BERT models help improve performance over baselines that do not use language model pretraining.

Besides multilingual models, language-specific BERT models have been trained for an increasing number of languages, including for instance CamemBERT \cite{Martin_2020} and  FlauBERT \cite{le2020flaubert} for French, FinBERT for Finnish \cite{virtanen2019multilingual}, RobBERT \cite{delobelle2020robbert} and BERTJe \cite{vries2019bertje} for Dutch, Chinese BERT \cite{cui2019pretraining}, BETO for Spanish \cite{canete2020spanish}, RuBERT for Russian \cite{kuratov2019adaptation} and others.
For a recent overview about these efforts refer to \citet{nozza2020mask}. Aggregating the results over different language-specific models and comparing them to those obtained with multilingual models shows that depending on the task, the average improvement of the language-specific BERT over the multilingual BERT varies from 0.70 accuracy points in paraphrase identification up to 6.37 in sentiment classification \cite{nozza2020mask}. The overall conclusion one can draw from these results is that while existing multilingual BERT models can bring along improvements over language-specific baselines, using language-specific BERT models can further considerably improve the performance of various NLP tasks.

Following the line of reasoning presented above, we set forth to train EstBERT, a language-specific BERT model for Estonian. In the following sections, we first give details about the data used for BERT pretraining and then describe the pretraining process. Finally, we will provide evaluation results on the same tasks as presented by \citet{kittask2020evaluating} on multilingual BERT models, which include POS and morphological tagging, named entity recognition and text classification. Additionally, we also train a dependency parser based on the spaCy system.
Compared to multilingual models, the EstBERT model achieves better results on five tasks out of seven, providing further evidence for the usefulness of pretraining language-specific BERT models.
Additionally, we also compare with the  Estonian WikiBERT, a recently published Estonian-specific BERT model trained on a relatively small Wikipedia data \citep{pyysalo2020wikibert}. Compared to the Estonian WikiBERT model, the EstBERT achieves better results on six tasks out of seven, demonstrating the positive effect of the amount of pretraining data on the generalisability of the model. 

\section{Data Preparation}

The first step for training the EstBERT model involves preparing a suitable unlabeled text corpus.
This section describes both the steps we took to clean and filter the data and the process of generating the vocabulary and the pretraining examples.

\subsection{Data Preprocessing}

For training the EstBERT model, we used the Estonian National Corpus 2017 \cite{national2017corpus},\footnote{\url{https://www.sketchengine.eu/estonian-national-corpus/}} which was the largest Estonian language corpus available at the time. It consists of four sub-corpora: the Estonian Reference Corpus 1990-2008, the Estonian Web Corpus 2013, the Estonian Web Corpus 2017, and the Estonian Wikipedia Corpus 2017. 
The Estonian Reference corpus (ca 242M words) consists of a selection of electronic textual material, about 75\% of the corpus contains newspaper texts, the rest 25\% contains fiction, science and legislation texts. The Estonian Web Corpora 2013 and 2017 make up the largest part of the material and they contain texts collected from the Internet. The Estonian Wikipedia Corpus 2017 is the Estonian Wikipedia dump downloaded in 2017 and contains roughly 38M words. 
The top row of the Table~\ref{tab:init_stat} shows the initial statistics of the corpus.


We applied different cleaning and filtering techniques to preprocess the data. 
First, we used the corpus processing methods
from EstNLTK \cite{laur-EtAl:2020:LREC}, which is an open-source tool for Estonian natural language processing. Using the EstNLTK, all XML/HTML tags were removed from the text, also all documents with a language tag other than Estonian were removed. Additional non-Estonian documents were further filtered out using the language-detection library.\footnote{\url{https://github.com/shuyo/language-detection}}
Next, all duplicate documents were removed. For that, we used hashing---all documents were lowercased, and then the hashed value of each document was subsequently stored into a set. Only those documents whose hash value did not yet exist in the set (i.e., the first document with each hash value) were retained. 
We also used the hand-written heuristics,\footnote{\url{https://github.com/TurkuNLP/deepfin-tools}} developed for preprocessing the data for training the FinBert model \cite{virtanen2019multilingual}, to filter out documents with too few words, too many stopwords or punctuation marks, for instance. We applied the same thresholds as were used for Finnish BERT.
Finally, the corpus was truecased by lemmatizing a copy of the corpus with EstNLTK tools and using the lemma's casing information to decide whether the word in the original corpus should be upper- or lowercase. 
The statistics of the corpus after the preprocessing and cleaning steps are in the bottom row of Table~\ref{tab:init_stat}.

\begin{table}[t]
\setlength{\tabcolsep}{5pt}
\centering
\begin{tabular}{lccc}
\toprule
                & \bf Documents & \bf Sentences & \bf Words \\
                \midrule
Initial  &   3.9M &   87.6M    &   1340M   \\ 
After cleanup  &    3.3M &   75.7M    &   1154M \\ 

\bottomrule
\end{tabular}
\caption{\label{tab:init_stat}Statistics of the corpus before and after the cleanup.}
\end{table}

\subsection{Vocabulary and Pretraining Example Generation}

Originally, BERT uses the WordPiece tokeniser, which is not available open-source. Instead, we used the BPE tokeniser available in the open-source sentencepiece\footnote{\url{https://github.com/google/sentencepiece}} library, which is the closest to the WordPiece algorithm, to construct a vocabulary of 50K subword tokens. Then, we used BERT tools\footnote{\url{https://github.com/google-research/bert}} to create the pretraining examples for the BERT model in the TFRecord format. 
In order to enable parallel training on four GPUs, the data was split into four shards.
Separate pretraining examples with sequences of length 128 and 512 were created, masking 15\% of the input words in both cases. Thus, 20 and 77 words in maximum were masked in sequences of both lengths, respectively.

\section{Evaluation Tasks}
\label{sec:tasks}

Before describing the EstBERT model pretraining process itself, we first describe the tasks used to both validate and evaluate our model. These tasks include the POS and morphological tagging, named entity recognition, and text classification. In the following subsection, we describe the available Estonian datasets for these tasks.

\subsection{Part of Speech and Morphological Tagging}
\label{subsec:POS_morph}

For part of speech (POS) and morphological tagging, we use the Estonian EDT treebank from the Universal Dependencies (UD) collection
that contains annotations of lemmas, parts of speech, universal morphological features, dependency heads, and universal dependency labels. We use the UD version 2.5 to enable comparison with the experimental results of the multilingual BERT models reported by \citet{kittask2020evaluating}.
We train models to predict both universal POS (UPOS) and language-specific POS (XPOS) tags as well as morphological features.  
The pre-defined train/dev/test splits are used for training and evaluation. Table~\ref{tab:ud_stats} shows the statistics of the treebank splits.  
The accuracy of the POS and morphological tagging tasks is evaluated with the \texttt{conll18\_ud\_eval} script from the CoNLL 2018 Shared Task.

\begin{table}[h]
\centering
\begin{tabular}{lrrr}
\toprule
\textbf{} & \textbf{Train} & \textbf{Dev} & \textbf{Test} \\
\midrule
 Sentences & 31012          & 3128         & 6348          \\
 Tokens  & 344646         & 42722        & 48491 \\
\bottomrule
\end{tabular}
\caption{\label{tab:ud_stats}Statistics of the UDv2.5 Estonian treebank.}
\end{table}

\subsection{Named Entity Recognition}

 Estonian named entity recognition (NER) corpus \cite{tkachenko2013named} annotations cover three types of named entities: locations, organizations, and persons. It contains 572 news stories published in local online newspapers Postimees and Delfi, covering local and international news on various topics. 
 Table~\ref{tab:ner_stats} displays statistics of the training, development and test splits. 
 The performance of the NER models is evaluated with the \texttt{conlleval} script from the CoNLL 2000 shared task.

\begin{table}[h!]
\centering
\begin{tabular}{lrrrrr}
\toprule
      & \bf Tokens & \bf{PER} & \bf{LOC} & \bf{ORG} & \bf{Total} \\
      \midrule
Train & 155981 & 6174         & 4749         & 4784         & 15707          \\
Dev   & 32890 & 1115          & 918          & 742          & 2775           \\
Test  & 28370 & 1201         & 644          & 619          & 2464 \\
\bottomrule
\end{tabular}
\caption{\label{tab:ner_stats}Statistics of the Estonian NER corpus.}
\end{table}

\subsection{Sentiment and Rubric Classification}

Estonian Valence corpus \cite{Pajupuu2016IDENTIFYINGPI} consists of 4085 news extracts from Postimees Daily. All documents in the corpus are labeled with both sentiment and rubric classes. There are nine rubrics: Opinion, Estonia, Life, Comments-Life, Comments-Estonia, Crime, Culture, Sports, and Abroad. The four sentiment labels include Positive, Negative, Neutral, and Ambiguous.
We split the data into 70/10/20 training, development and test sets, stratified over both rubric and sentiment analysis.
Table~\ref{tab:sentiment_stats} and Table~\ref{tab:text_stats} show the statistics about the sentiment and rubric view of the classification dataset respectively. 

\begin{table}[t]
\centering
\begin{tabular}{lrrrr}
\toprule
 & \bf Train    & \bf Dev   & \bf Test  & \bf Total \\
\midrule
Positive    & 612   & 87    & 175   & 874 \\
Negative    & 1347  & 191   & 385   & 1923 \\
Neutral     & 505   & 74    & 142   & 721 \\
Ambiguous   & 385   & 55    & 110   & 550 \\
\midrule
Total       & 2849  & 407   & 812   & 4068 \\
\bottomrule
\end{tabular}
\caption{\label{tab:sentiment_stats}Sentiment label statistics of the Estonian Valence corpus.}
\end{table}

\begin{table}[t]
\centering
\begin{tabular}{lrrrr}
\toprule
 & \bf Train & \bf Dev   & \bf Test  & \bf Total \\
\midrule
Opinion             & 676   & 96    & 192   & 964 \\
Estonia             & 289   & 41    & 83    & 413 \\
Life                & 364   & 52    & 101   & 517 \\
Comments-Life       & 354   & 50    & 102   & 506 \\
Comments-Estonia    & 351   & 50    & 100   & 501 \\
Crime               & 146   & 21    & 42    & 209 \\
Culture             & 182   & 27    & 51    & 260 \\
Sports              & 269   & 39    & 77    & 385 \\
Abroad              & 218   & 31    & 64    & 313 \\
\midrule
Total               & 2849  & 407   & 812   & 4068 \\
\bottomrule
\end{tabular}
\caption{\label{tab:text_stats}Rubric label statistics of the Estonian Valence corpus.}
\end{table}

\section{Pretraining EstBERT}

\begin{figure*}[t]
\centering
\begin{subfigure}{0.49\linewidth}
    \includegraphics[width=\linewidth]{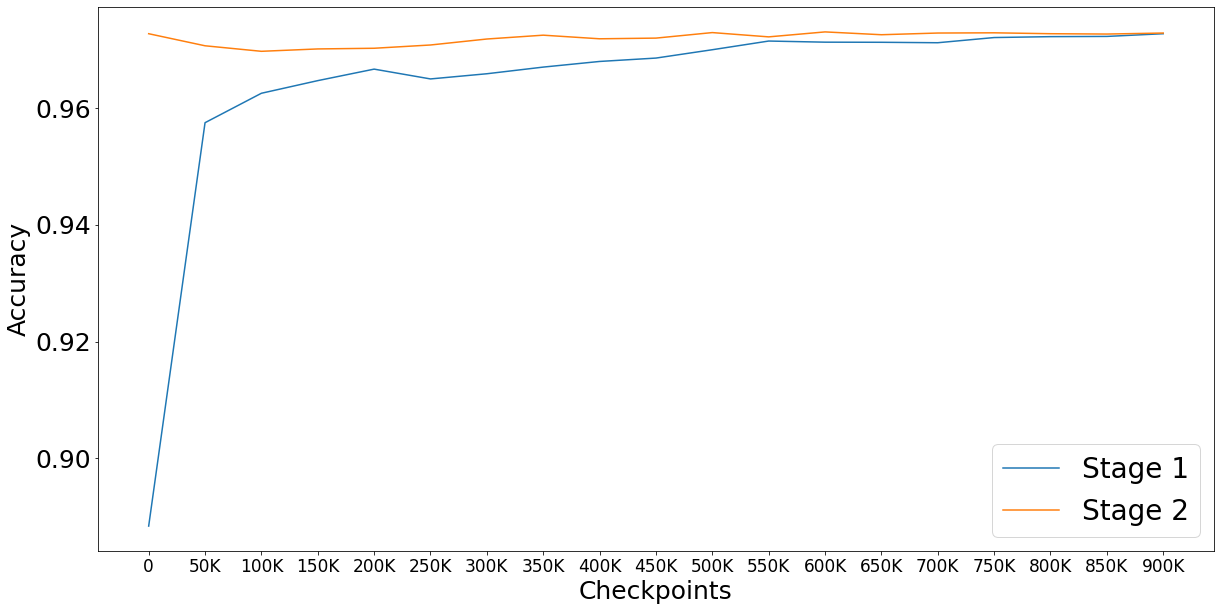}
    \caption{Universal POS tags}
    \label{fig:short_UPOS}
\end{subfigure}
\begin{subfigure}{0.49\linewidth}
    \includegraphics[width=\linewidth]{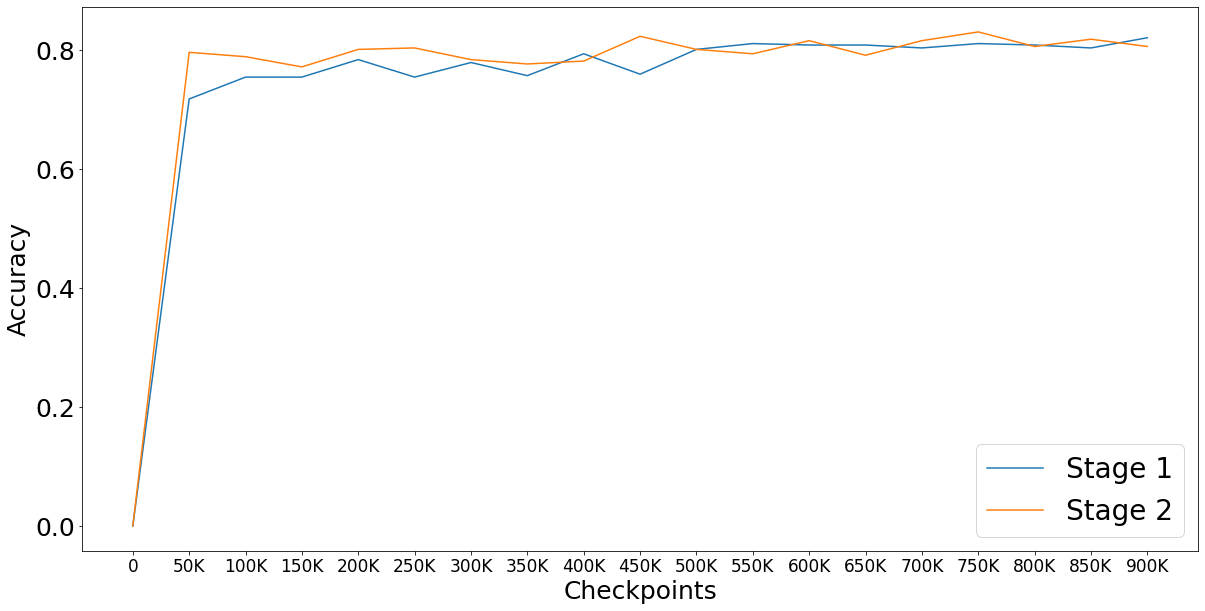}
    \caption{Rubric Classification}
    \label{fig:short_TOPIC}
\end{subfigure}
\caption{The validation performance on POS tagging and text classification tasks after every 50K checkpoints.}\label{fig:short_val}
\end{figure*} 

The EstBERT model was pretrained on the architecture identical to the BERT\textsubscript{Base} model with 12 transformer blocks with 768 hidden units each and 110M trainable parameters. It was pretrained on the Masked Language Modeling (MLM) and the Next Sentence Prediction (NSP) tasks as described by \citet{devlin2019bert}. 
In MLM, the probability of correctly predicting the randomly masked tokens is maximised. Because in the transformer architecture, the model can simultaneously see both the left and the right context of a masked word, optimizing the MLM gives the model a bidirectional understanding of a sentence, as opposed to only the left or right context provided by recurrent language models.
 The NSP involves optimizing a binary classification task to predict whether the two sequences in the input follow each other in the original text or not, where half of the time, the second sequence is the correct next sentence and the other half of the time the two sequences are unrelated.  
The models were trained on four NVIDIA Tesla V100 GPUs across two nodes of the High-performance Computing Center at the University of Tartu \cite{HPC}.

The model was first trained with the sequence length of 128. Then we evaluated the checkpoints generated during pretraining on the tasks described in section~\ref{sec:tasks} to choose the final model with that sequence length. Finally, the chosen model was used as a starting point for training the longer model with 512 sequence length.
Thus, as a result of pretraining, two EstBERT models, one with maximum sequence length 128 and the other with maximum sequence length 512, were obtained.
The following subsections describe these three steps in more detail.

\subsection{Pretraining with Sequence Length 128}

The model with the sequence length of 128 was pretrained for two phases, both for 900K steps. Although the number of training steps was chosen following \citet{virtanen2019multilingual}, coincidentally this (900K steps) was the maximum number of steps we could fit in the given GPU time limit of 8 days. Therefore, the model was trained in two phases, each having 900K steps. 
A checkpoint was saved to the disk after every 50K steps.
While the first phase of pretraining started from scratch with randomly initialised parameters, the second phase of training was initialised from the first phase's last checkpoint.
Since the GPU memory availability was a major issue, the batch size was kept at 32 to avoid the tensors going beyond the allowed GPU memory size.
The BERT\textsubscript{Base} uses Adam optimiser with weight decay. For EstBERT, the same optimiser was used with warmup over the first 1\% of steps (9000) to a peak learning rate of 1e-4. 
The relevant hyperparameters are shown in Table~\ref{tab:short_params}.
The pretraining process took around 192 hours for each phase. 

\begin{table}
    \centering
    \begin{tabular}{ll}
    \toprule
       train\_batch\_size  & 32 \\
        max\_seq\_length & 128 \\
        max\_predictions\_per\_seq & 20 \\
        num\_train\_steps & 900000 \\
        num\_warmup\_steps & 9000 \\
        learning\_rate & 1e-4 \\
        save\_checkpoints\_step & 50000 \\
    \bottomrule
    \end{tabular}
    \caption{Hyperparameters used in the first two pretraining phases with the sequence length 128.}
    \label{tab:short_params}
\end{table}

\subsection{Pretraining Validation}
\label{sec:validation}

During pretraining, a checkpoint was saved after every 50K steps for later evaluation. 
To monitor the pretraining process, we evaluated the performance of MLM, NSP, and the evaluation tasks described in section~\ref{sec:tasks} on all these checkpoints.

For POS and morphological tagging, and named entity recognition, we finetuned EstBERT using scripts from HuggingFace transformers library.\footnote{\url{https://github.com/huggingface/transformers/blob/master/examples/token-classification/run_ner.py}} 
A single randomly initialised fully connected classifier layer was trained on top of the token representations of the last hidden layer of the EstBERT model.
All hyperparameters were kept at their default values, which involves training for three epochs, using the learning rate of 5e-5 and batch size of 8.
For the rubric and sentiment classification tasks, we adapted the classifier training scripts available in google's BERT repository.\footnote{\url{https://github.com/google-research/bert/blob/master/run_classifier.py/}}
The input to the single fully-connected classifier layer is the last hidden representation of the first token [CLS] in the input sequence.
Here again, the classifier layer was initialised randomly and the default values for hyperparameters were used: training for three epochs with the learning rate 5e-5 and batch size 32.
In all tasks, both the task-specific classification layer as well as the EstBERT parameters were finetuned. 

The validation results of the masked language model, next sentence prediction accuracy, and all the evaluation tasks for all the eighteen checkpoints from stage one and other eighteen models from stage two were compared to pick the best model. The examples of validation curves for the UPOS tagging and the rubric classification tasks are shown in Figure \ref{fig:short_val}.
 Although the checkpoint validation results from both phases showed more or less steady improvement with the increase of the number of steps trained, we observed that the checkpoint at 750K steps from phase two performs slightly better on all tasks than the rest of the checkpoints.
Thus, this checkpoint was chosen as a final model with sequence length 128.

\begin{table}[th]
    \centering
    \begin{tabular}{ll}
    \toprule
       train\_batch\_size  & 16 \\
        max\_seq\_length & 512 \\
        max\_predictions\_per\_seq & 77 \\
        num\_train\_steps & 600000 \\
        num\_warmup\_steps & 6000 \\
        learning\_rate & 1e-4 \\
        save\_checkpoints\_step & 50000 \\
    \bottomrule
    \end{tabular}
    \caption{Hyperparameters used to pretrain the EstBERT model with the sequence length 512.}
    \label{tab:long_params}
\end{table}

\begin{table*}[ht]
\centering
\begin{tabular}{lcccccc}
\toprule
 \bf Model & \bf UPOS  & \bf XPOS & \bf Morph & \bf UPOS  & \bf XPOS & \bf Morph \\
 & \multicolumn{3}{c}{\bf Seq = 128} & \multicolumn{3}{c}{\bf Seq = 512} \\
\midrule
EstBERT            & \bf \underline{97.89} & \bf 98.40 & \bf 96.93 &\bf 97.84 & \bf \underline{98.43} &  \bf \underline{96.80}    \\
\midrule
WikiBERT-et & 97.78 & 98.36 & 96.71 & 97.76 & 98.35 & 96.67 \\
mBERT                  & 97.42 & 98.06 & 96.24 & 97.43 & 98.13 & 96.13  \\
XLM-RoBERTa            & 97.78 & 98.36 & 96.53 & 97.80 & 98.40 &  96.69    \\

\bottomrule
\end{tabular}
\caption{\label{tab:pos_results}POS and morphological tagging accuracy on the Estonian UD test set. The highest scores in each column are in \textbf{bold}. The highest overall score of each task is \underline{underlined}.}
\end{table*}

\subsection{Pretraining with Sequence Length 512}

The starting point for training the model with a sequence length of 512 was the final model chosen for the sequence length 128.
 The longer model was trained further up to 600K steps.
The batch size was reduced to 16 as the size of the tensors would be larger for the sequence length 512 compared to 128. 
The hyperparameters used to train the longer model are shown in Table~\ref{tab:long_params}.
During training, checkpoints were again saved after every 50K steps, and these were evaluated on all evaluation tasks as previously explained in Section~\ref{sec:validation}. Based on these evaluations, the last checkpoint obtained after the 600K steps was chosen as the final model with 512 sequence length. 

\section{Results}

The next subsections present the results obtained with the final EstBERT models with both sequence lengths on the tasks described in section~\ref{sec:tasks}.
We follow the same setup of \citet{kittask2020evaluating} to enable direct comparison with the multilingual models. 
Some additional steps were taken to prepare the Estonian Valence corpus. First, all duplicate items, 17 in total, were removed. Also, all items with the Ambiguous label were removed as retaining them has been shown to lower the the classification accuracy \cite{Pajupuu2016IDENTIFYINGPI}. 
The same preprocessing was also applied in evaluating the multilingual BERT models for Estonian \cite{kittask2020evaluating}. 

For finetuning, we used the same scripts from the HuggingFace transformers repository that were used for the pretraining validation in section \ref{sec:validation}. The same scripts were also used to evaluate the multilingual models by \citet{kittask2020evaluating}. 
For each task, the learning rate of the AdamW optimiser and the batch size was tuned on the development set.  The learning rate grid values were (5e-5, 3e-5, 1e-5, 5e-6, 3e-6) and the batch size grid values were (8, 16). The best model was found on the development set by using early stopping with the patience of 10 epochs. 

We compare the results of EstBERT with the multilingual BERT models' results from \citet{kittask2020evaluating} and the WikiBERT model trained on the Estonian Wikipedia \cite{pyysalo2020wikibert}. WikiBERT-et model was finetuned using the same setup described above.

\subsection{POS and Morphological Tagging}

The POS and morphological tagging results are summarised in Table~\ref{tab:pos_results} that shows the accuracy for universal POS tags (UPOS), language-specific POS tags (XPOS), and morphological features.
The language-specific EstBERT outperforms all other models although the difference with the XLM-RoBERTa---the best-performing multilingual model---and the WikiBERT-et are quite small. 

Similar to multilingual results, using longer sequence length on this task with the EstBERT model does not seem beneficial as the accuracy slightly increases only for XPOS tags but not for others.  Overall, as the performances on these tasks are already very high, the absolute performance gains cannot be large. 
EstBERT obtains consistent improvements over mBERT,  with the relative error reduction with both models on all tasks falling between 16-18\%. The relative error reduction of the EstBERT compared to XLM-RoBERTa is smaller,  in the range of 2-5\%. The highest reduction of error of EstBERT compared to XLM-RoBERTa can be observed on the morphological tagging task with the shorter model where the relative error reduction is 12\%. The WikiBERT-et model achieves almost identical results to XLM-RoBERTa with both sequence lengths.

\subsection{Rubric and Sentiment Classification}

\begin{table}[t]
\centering
\begin{tabular}{lccll}
\toprule
\textbf{Model} & \textbf{Rubr.}      & \textbf{Sent.}   & \multicolumn{1}{c}{\textbf{Rubr.}} & \textbf{Sent.} \\
               & \multicolumn{2}{c}{\textbf{Seq = 128}}      & \multicolumn{2}{c}{\textbf{Seq = 512}} \\
\midrule
EstBERT    &   \bf \underline{81.70}     &  74.36  &   \bf 80.96   &  74.50          \\
\midrule
WikiBERT-et & 72.72 & 68.09 & 71.13 & 69.37 \\ 
mBERT       &    75.67     &  70.23  & 74.94  &  69.52  \\
XLM-RoBERTa &    80.34     & \bf 74.50  & 78.62   &  \bf \underline{76.07}          \\

\bottomrule
\end{tabular}
\caption{\label{tab:sent_results} Rubric (Rubr.) and sentiment (Sent.) classification accuracy. The highest scores in each column are in \textbf{bold}. The highest overall score of each task is \underline{underlined}.}
\end{table}

\begin{table*}[t]
\centering
\begin{tabular}{lcccccc}
\toprule
 \multirow{2}{*}{\bf Model} & \bf Precicion  & \bf Recall & \bf F1-Score & \bf Precision  & \bf Recall & \bf F1-Score  \\
 & \multicolumn{3}{c}{\bf Seq = 128} & \multicolumn{3}{c}{\bf Seq = 512} \\
\midrule
EstBERT            &  89.10  &  91.15 &   90.11 & 88.35 & 89.74   &   89.04 \\
\midrule
WikiBERT-et & \bf \underline{89.86} &  90.83 & \bf \underline{90.34} &  88.31 & \bf 90.96 & \bf 89.61 \\
mBERT                  & 85.88 & 87.09 & 86.51 & \bf 88.47 & 88.28 & 88.37 \\

XLM-RoBERTa            & 87.55 &   \bf \underline{91.19} &  89.34 & 87.50 &  90.76 &  89.10 \\

\bottomrule
\end{tabular}
\caption{\label{tab:ner_results}NER tagging results. Upper section shows the comparison between different models. The highest scores in each column are in \textbf{bold}. The highest overall score of each measure is \underline{underlined}.}
\end{table*}

\begin{table*}[t]
\centering
\begin{tabular}{lccccccccc}
\toprule
\multirow{2}{*}{\bf Entity}& \multicolumn{3}{c}{\bf EstBERT} & \multicolumn{3}{c}{\bf XLM-RoBERTa} & \multicolumn{3}{c}{\bf WikiBERT-et} \\
 & \bf Prec  & \bf Rec & \bf F1 & \bf Prec  & \bf Rec & \bf F1 & \bf Prec  & \bf Rec & \bf F1 \\
\midrule
PER & 94.80 & \bf 95.77 & 95.28 & \bf 96.42 & 94.45 & \bf 95.43 & 94.87	& 94.45	& 94.66\\
ORG & 78.38 & \bf 82.64 & 80.45 & 75.48 & 82.12 & 78.66 & \bf 82.25	& 81.61	& \bf 81.92\\
LOC & \bf 89.94 & 91.38 & 90.66 & 86.06 & \bf 93.99 & 89.85 & 88.89	& 92.99	& \bf 90.89\\
\bottomrule
\end{tabular}
\caption{\label{tab:entity_results}The entity-based scores for the EstBERT, XLM-RoBERTa and the WikiBERT-et models. The best scores are in \textbf{bold}.}
\end{table*}

The rubric and sentiment classification results are shown in Table~\ref{tab:sent_results}. EstBERT outperforms mBERT and WikiBERT-et on both tasks by a large margin, but XLM-RoBERTa exceeds EstBERT on sentiment classification. The difference between the two accuracy scores is relatively small when the model with sequence length 128 is used, but it increases when the longer sequence length is used. 

Like XLM-RoBERTa, the EstBERT model with a shorter sequence length is somewhat better on rubric classification, and the opposite is true for sentiment classification. Overall, the differences between the EstBERT models' performances with both sequence lengths are again relatively small.

\subsection{Named Entity Recognition}

Table~\ref{tab:ner_results} shows the entity-based precision, recall, and F-score of the named entity recognition task. 
WikiBERT-et model is the best model in this task, obtaining the highest F1-score with both the short and long models and the overall highest F1-score with the short model.
XLM-RoBERTa achieves the highest recall in the short model category but remains below the EstBERT in terms of the F1-score.
EstBERT, WikiBERT-et and XLM-RoBERTa all benefit from using the smaller sequence length rather than longer, while mBERT shows the opposite behavior.


Table~\ref{tab:entity_results} shows the fine-grained scores of each entity type for both the EstBERT,  XLM-RoBERTa and the WikiBERT-et shorter model. In alignment with the previous results in Estonian NER \cite{tkachenko2013named}, the prediction of the person entities is the most accurate while the organisation names are the most difficult to predict. The WikiBERT-et is the best on the two most difficult entities ORG and LOC, while the EstBERT model is better than XLM-RoBERTa on these two entities. 
The WikiBERT-et is notably the best on the most challenging organisation entity, improving the precision over the EstBERT model for almost 4\% and over the XLM-RoBERTa for almost 7\%, with a considerably smaller loss in recall. One reason for the superiority of the WikiBERT-et model might lie in the fact that the Wikipedia dataset used to train the WikiBERT-et model probably contains a much higher proportion of organisation names. Although the datasets used to train the other two models also contain the Estonian Wikipedia dataset, it has been diluted in other languages (in case of XLM-RoBERTa) or genres (in case of EstBERT). However, this is just a hypothesis at the moment that has to be studied more in further work.


\subsection{Dependency Parsing}

\begin{table*}[t]
\begin{small}
    \setlength\tabcolsep{5pt}
    \centering
    \begin{tabular}{l@{\hspace{0.2em}}r@{\hspace{0.4em}}rrrrrrrrrrrr}
    \toprule
        \bf Model &  & \multicolumn{3}{c}{\bf EstBERT} & \multicolumn{3}{c}{\bf XLM-RoBERTa} & \multicolumn{3}{c}{\bf WikiBERT-et} & \multicolumn{3}{c}{\bf Stanza}\\
     \bf DepRel & \bf Support & \bf Prec & \bf Rec & \bf F1 & \bf Prec & \bf Rec & \bf F1 & \bf Prec & \bf Rec & \bf F1 & \bf Prec & \bf Rec & \bf F1 \\
    \midrule
        \bf UAS & & 86.07 & \underline{87.34} & \underline{86.70} & \bf 88.02 & \bf 89.32 & \bf 88.66 & 85.97 & 87.24 & 86.60 &\underline{86.69} & 86.68 & 86.69\\
        \bf LAS & & 83.32 & \underline{84.56} & \underline{83.94} & \bf 85.60 & \bf 86.87 & \bf 86.23 & 83.06 & 84.29 & 83.67 & \underline{83.63} & 83.63 & 83.63\\
   
    \midrule
nmod & 4328 & 81.40 & \underline{85.65} & \underline{83.47} & \bf 83.73 & \bf 88.49 & \bf 86.05 & 80.86 & 85.42 & 83.08 & \underline{82.53} & 84.27 & 83.39 \\
obl & 4198 & \underline{80.99} & \underline{79.78} & \underline{80.38} & \bf 83.65 & \bf 82.66 & \bf 83.15 & 80.81 & 79.47 & 80.13 & 79.61 & 77.78 & 78.68 \\
advmod & 3938 & 78.01 & \underline{79.02} & \underline{78.52} & \bf 80.43 & \bf 80.45 & \bf 80.44 & 78.08 & 77.96 & 78.02 & \underline{78.93} & 78.11 & \underline{78.52} \\
root & 3214 & \underline{90.82} & 89.61 & \underline{90.21} & \bf 91.88 & \bf 91.91 & \bf 91.90 & 90.32 & \underline{89.73} & 90.03 & 90.18 & 87.46 & 88.80 \\
nsubj & 2682 & \underline{92.05} & \underline{93.25} & \underline{92.65} & \bf 93.54 & \bf 94.52 & \bf 94.03 & 90.40 & 92.69 & 91.53 & 90.67 & 89.90 & 90.28 \\
conj & 2476 & 76.90 & 78.51 & 77.70 & \bf 81.72 & \bf 83.44 & \bf 82.57 & \underline{78.28} & 78.31 & \underline{78.30} & 76.41 & \underline{78.76} & 77.57 \\
obj & 2437 & \underline{86.91} & \underline{88.80} & \underline{87.84} & \bf 88.62 & \bf 90.73 & \bf 89.66 & 86.36 & 87.57 & 86.96 & 83.51 & 84.78 & 84.14 \\
amod & 2411 & 80.02 & 84.20 & 82.05 & 82.90 & 86.64 & 84.73 & 80.12 & 83.91 & 81.97 & \bf \underline{91.93} & \bf \underline{89.26} & \bf \underline{90.57} \\
cc & 2029 & \underline{91.26} & \underline{90.09} & \underline{90.67} & \bf 92.57 & \bf 91.47 & \bf 92.02 & 90.75 & 89.50 & 90.12 & 89.97 & 88.42 & 89.19 \\
aux & 1372 & \underline{95.36} & \underline{95.85} & \underline{95.60} & \bf 95.43 & \bf 95.99 & \bf 95.71 & 94.79 & 95.48 & 95.13 & 89.93 & 95.04 & 92.42 \\
mark & 1277 & \underline{90.14} & \underline{90.92} & \underline{90.53} & \bf 92.75 & \bf 93.19 & \bf 92.97 & 89.25 & 89.74 & 89.50 & 88.35 & 89.12 & 88.73 \\
cop & 1202 & \underline{84.75} & \underline{87.35} & \underline{86.03} & \bf 85.48 & \bf 87.69 & \bf 86.57 & 84.29 & 87.02 & 85.63 &  81.43 & 86.11 & 83.70 \\
acl & 1063 & 84.98 & \underline{85.14} & 85.06 & \bf 86.88 & \bf 87.86 & \bf 87.37 & \underline{86.67} & 85.04 & \underline{85.85} & 86.36 & 80.43 & 83.29 \\
nsubj:cop & 1054 & \underline{79.98} & 82.64 & \underline{81.29} & \bf 81.16 & \bf 85.01 & \bf 83.04 & 79.34 & \underline{82.73} & 81.00 & 77.78 & 79.70 & 78.73 \\
case & 908 & \underline{92.42} & 92.62 & \underline{92.52} & \bf 93.52 & \bf 93.72 & \bf 93.62 & 91.32 & \underline{92.73} & 92.02 & 89.13 & 91.19 & 90.15 \\
advcl & 857 & \underline{67.18} & 65.93 & \underline{66.55} & \bf 73.46 & \bf 71.06 & \bf 72.24 & 66.55 & \underline{66.39} & 66.47 & 67.12 & 63.36 & 65.19 \\
det & 808 & \underline{83.88} & \underline{85.02} & \underline{84.45} & \bf 87.89 & \bf 87.13 & \bf 87.51 & 82.95 & 84.28 & 83.61 & 80.80 & 82.80 & 81.78 \\
parataxis & 725 & 52.96 & 49.38 & 51.11 & 57.50 & 50.76 & 53.92 & 55.59 & 48.69 & 51.91 & \bf \underline{65.45} & \bf \underline{59.31} & \bf \underline{62.23} \\
xcomp & 641 & \underline{85.21} & \underline{87.21} & \underline{86.20} & \bf 88.06 & \bf 88.61 & \bf 88.34 & 84.11 & 86.74 & 85.41 & 83.78 & 83.00 & 83.39 \\
flat & 633 & 81.44 & 85.94 & 83.63 & 86.64 & 91.15 & 88.84 & 80.09 & 86.41 & 83.13 &\bf \underline{88.60} & \bf \underline{92.10} & \bf \underline{90.32} \\
nummod & 555 & 62.88 & 77.84 & 69.57 & 62.00 & 80.54 & 70.06 & 63.12 & 78.02 & 69.78 & \bf \underline{85.53} & \bf \underline{85.23} & \bf \underline{85.38} \\
compound:prt & 481 & \underline{86.10} & 92.72 & 89.29 & \bf 88.20 & \bf 94.80 & \bf 91.38 & 85.99 & \underline{93.14} & \underline{89.42} & 85.52 & 89.60 & 87.51 \\
appos & 376 & 69.07 & 71.28 & 70.16 & \bf 74.45 & \bf 80.59 & \bf 77.39 & 64.55 & \underline{73.14} & 68.58 & \underline{69.47} & 72.61 & \underline{71.00} \\
ccomp & 344 & \underline{82.56} & 82.56 & 82.56 & \bf 87.03 & \bf 87.79 & \bf 87.41 & 80.44 & \underline{84.88} & \underline{82.60} & 81.87 & 78.78 & 80.30 \\
acl:relcl & 315 & \bf \underline{80.67} & \bf \underline{83.49} & \bf \underline{82.06} & 79.00 & 80.00 & 79.50 & 79.30 & 79.05 & 79.17 & 61.32 & 82.54 & 70.37 \\
csubj:cop & 121 & \bf \underline{80.47} & 85.12 & \bf \underline{82.73} & 75.74 & 85.12 & 80.16 & 79.23 & 85.12 & 82.07 & 72.79 & \bf \underline{88.43} & 79.85 \\
csubj & 108 & 81.51 & \bf \underline{89.81} & \bf \underline{85.46} & 80.83 & \bf 89.81 & 85.09 & 84.26 & 84.26 & 84.26 & \bf \underline{84.91} & 83.33 & 84.11 \\
discourse & 47 & 37.14 & \bf \underline{55.32} & 44.44 & 36.92 & 51.06 & 42.86 & 34.33 & 48.94 & 40.35 & \bf \underline{81.25} & \bf \underline{55.32} & \bf \underline{65.82} \\
orphan & 44 & 20.83 & 11.36 & 14.71 & 37.93 & \bf 25.00 & \bf 30.14 & 20.00 & 18.18 & 19.05 & \bf \underline{45.00} & \underline{20.45} & \underline{28.12} \\
compound & 43 & 88.10 & 86.05 & 87.06 & 83.33 & 81.40 & 82.35 & \bf \underline{92.11} & 81.40 & 86.42 & 88.64 & \bf \underline{90.70} & \bf \underline{89.66} \\
cc:preconj & 39 & 66.67 & \bf \underline{71.79} & \bf \underline{69.14} & 67.57 & 64.10 & 65.79 & 62.79 & 69.23 & 65.85 & \bf \underline{70.27} & 66.67 & 68.42 \\
flat:foreign & 37 & \underline{76.19} & 43.24 & \underline{55.17} & \bf 87.50 & \bf 56.76 & \bf 68.85 & 43.75 & 18.92 & 26.42 & 65.38 & \underline{45.95} & 53.97 \\
fixed & 31 & 64.71 & 70.97 & 67.69 & 64.86 & \bf 77.42 & 70.59 & 57.50 & \underline{74.19} & 64.79 & \bf \underline{75.86} & 70.97 & \bf \underline{73.33} \\
vocative & 9 & 22.22 & 22.22 & 22.22 & 28.57 & 22.22 & 25.00 & 5.56 & 11.11 & 7.41 & \bf \underline{30.77} & \bf \underline{44.44} & \bf \underline{36.36} \\
goeswith & 8 & \bf \underline{100.00} & 12.50 & 22.22 & 25.00 & 12.50 & 16.67 & 33.33 & \bf \underline{25.00} & \bf \underline{28.57} & 0.00 & 0.00 & 0.00 \\
dep & 5 & 0.00 & 0.00 & 0.00 & 0.00 & 0.00 & 0.00 & 0.00 & 0.00 & 0.00 & 0.00 & 0.00 & 0.00 \\
list & 1 & 0.00 & 0.00 & 0.00 & 0.00 & 0.00 & 0.00 & 0.00 & 0.00 & 0.00 & 0.00 & 0.00 & 0.00 \\
\bottomrule
    \end{tabular}
    \caption{Dependency parsing results. The best scores over all models are in \textbf{bold}. The best scores comparing the EstBERT, WikiBERT-et and the Stanza models are \underline{underlined}.}
    \label{tab:parsing_results}
\end{small}
\end{table*}

Additionally, we also evaluated both the EstBERT, WikiBERT-et and the XLM-RoBERTa models on the Estonian dependency parsing task. The data used in these experiments is the Estonian UDv2.5 described in section \ref{subsec:POS_morph}. We trained the parser available in the spaCy Nightly version\footnote{\url{https://pypi.org/project/spacy-nightly/}} that also supports transformers. The models were trained with a batch size of 32 and for a maximum of 20K steps, stopping early when the development set performance did not improve for 1600 steps. The parser was trained jointly with a tagger component that predicted the concatenation of POS tags and morphological features. During training, the model was supplied with the gold sentence and token segmentations. During evaluation, the sentence segmentation and tokenisation was done with the out-of-the-box spaCy tokeniser.

The dependency parsing results are in Table \ref{tab:parsing_results}. In addition to the transformer-based EstBERT and XLM-RoBERTa models, the right-most section also displays the Stanza parser \cite{qi-etal-2020-stanza}, trained on the same Estonian UDv2.5 corpus, obtained from the Stanza web page.\footnote{\url{https://stanfordnlp.github.io/stanza/models.html}} We add a non-transformer based baseline for this task because dependency parsing was not evaluated by \citet{kittask2020evaluating}. Overall, the XLM-RoBERTa model performs the best, both in terms of the UAS and LAS metrics and the individual dependency relations. This is especially true for dependency relations with larger support in the test set. Although in terms of the UAS and LAS, the EstBERT, WikiBERT-et and Stanza models seem to perform similarly, a closer look into the scores of the individual dependency relations reveals that in most cases, especially with relations of larger support, the EstBERT model performs the best. There are few dependency relations where the Stanza system's predictions are considerably more accurate than the BERT-based models, the most notable of them being the adjectival modifier (amod) and the numerical modifier (nummod). Further analyses are needed to gain more insight into these results.

\section{Discussion}

This objective of this paper was to describe the process of pretraining the language-specific BERT model for Estonian and to compare its performance with the multilingual BERT models as well as with the smaller Estonian WikiBERT model on several NLP tasks.
Overall, the pretrained EstBERT was better than the best multilingual XLM-RoBERTa model on five tasks out of seven: UPOS, XPOS, and morphological tagging, rubric classification, and NER. Only in the sentiment classification and dependency parsing tasks, the XLM-RoBERTa was better. 
Compared to WikiBERT-et, the EstBERT model was better on six tasks out of seven---the WikiBERT-et model was superior  only in the NER task, predicting ORG entities considerably better than any other model.
We did not observe any consistent difference between the models of different sequence lengths, although the model with the sequence length 512 was trained longer. It is possible that the shorter model was already trained long enough, and the subsequent training of the longer model did not add any effect in that respect, aside from the fact that it can accept longer input sequences.

One crucial aspect of this work was obtaining a large-enough corpus for pretraining the model. We used the Estonian National Corpus 2017, which was the largest corpus available at the time. A newer and larger version of this corpus---the Estonian National Corpus 2019 \cite{national2019corpus}---has become available meanwhile. There are also few other resources, such as the Estonian part of the CoNLL 2017 raw data \cite{conll2017corpus} and the Oscar Crawl, which probably partially overlap with each other and with the Estonian National Corpus. Still, these corpora would potentially provide additional data that was currently not used.

Another challenge was related to finding annotated datasets for downstream tasks. While the Estonian UD dataset provides annotations to the common dependency parsing pipeline tasks, datasets for other, especially semantic NLP tasks, are scarce. We adopted the Estonian Valence corpus for two-way text classification. However, the labels of this corpus are semi-automatically derived from user ratings, and thus the quality of these annotations cannot be guaranteed. An Estonian coreference dataset with some simple baseline results in nominal coreference resolution has recently become available \cite{barbu2020a}, which gives further opportunities to test out the EstBERT model in future work.

When preprocessing the data and pretraining the model, we mostly followed the process of training the FinBERT model for Finnish \cite{virtanen2019multilingual}. We also decided to truecase our corpus to reduce the number of capitalised words in the vocabulary.
The EstBERT model itself was also pretrained on the truecased corpus. However, when training the task-based models for evaluation, the EstBERT was finetuned on the cased datasets. Thus, truecasing the datasets before finetuning might have a positive effect on the results. In order to verify this, the EstBERT-based task-specific models should be finetuned using the truecased annotated datasets as input, and compared with the results reported in this paper. 

Although we did see some improvements with EstBERT compared to XLM-RoBERTa on the smaller model for the NER task, the differences in scores were generally relatively small. 
However, we have observed that the annotations of this NER dataset are occasionally erroneous, containing, for instance, label sequences (I-PER, I-PER) instead of (B-PER, I-PER). We have also observed unlabelled entities in the text. Thus, the small variations in the systems' results might not be informative about the systems themselves but can instead stem from the noise in the data. Although these annotation errors have been noticeable enough, the magnitude of these errors has not been quantified.

The differences between the EstBERT and the XLM-RoBERTa model were, in most cases, relatively small. In previous experiments with several multilingual BERT models on the same Estonian tasks \cite{kittask2020evaluating}, the XLM-RoBERTa proved to be the best multilingual model. This suggests that one option to obtain an even better Estonian language-specific model would be to train an Estonian-specific RoBERTa by initializing the model with the parameters of the XLM-RoBERTa. Considering that the multilingual RoBERTa already performs very well on Estonian tasks, finetuning it with more Estonian data would hopefully bias it even more to the Estonian language while also maintaining the gains obtained from multilingualism.

\section{Conclusion}

We presented EstBERT, the largest BERT model pretrained specifically on the Estonian language. While several existing multilingual BERT models include Estonian, the only language-specific Estonian BERT model available until now has been trained on the relatively small Wikipedia data. In order to pretrain the EstBERT model, we used the largest Estonian text corpus available at the time. The EstBERT model was put to the test by finetuning it for several tasks, including POS and morphological annotations, dependency parsing, named entity recognition, and text classification. On five tasks out of seven, the classifiers based on EstBERT achieved better performance than the models based on multilingual BERT models, although in several cases, the gap with the best-performing multilingual XLM-RoBERTa was relatively small. These results suggest that training a RoBERTa model for Estonian, initialised with the multilingual model's parameters, might be beneficial. On six tasks out of seven, the models based on EstBERT achieved better results than the Estonian BERT model trained on Wikipedia, suggesting that using more textual data for pretraining leads to a more generalisable model.

\bibliographystyle{acl_natbib}
\bibliography{bibliography}

\begin{thebibliography}{25}
\expandafter\ifx\csname natexlab\endcsname\relax\def\natexlab#1{#1}\fi

\bibitem[{Barbu et~al.(2020)Barbu, Muischnek, and Freienthal}]{barbu2020a}
Eduard Barbu, Kadri Muischnek, and Linda Freienthal. 2020.
\newblock \href {https://doi.org/10.3233/FAIA200595} {{A Study in Estonian
  Pronominal Coreference Resolution}}.
\newblock In \emph{Volume 328: Human Language Technologies – The Baltic
  Perspective}, Frontiers in Artificial Intelligence and Applications, pages
  3--10.

\bibitem[{Cañete et~al.(2020)Cañete, Chaperon, Fuentes, and
  P{\'e}rez}]{canete2020spanish}
Jos{\'e} Cañete, Gabriel Chaperon, Rodrigo Fuentes, and Jorge P{\'e}rez. 2020.
\newblock {Spanish pre-trained BERT model and evaluation data}.
\newblock \emph{PML4DC at ICLR}.

\bibitem[{Conneau et~al.(2019)Conneau, Khandelwal, Goyal, Chaudhary, Wenzek,
  Guzm{\'a}n, Grave, Ott, Zettlemoyer, and Stoyanov}]{conneau2019unsupervised}
Alexis Conneau, Kartikay Khandelwal, Naman Goyal, Vishrav Chaudhary, Guillaume
  Wenzek, Francisco Guzm{\'a}n, Edouard Grave, Myle Ott, Luke Zettlemoyer, and
  Veselin Stoyanov. 2019.
\newblock \href {https://arxiv.org/abs/1911.02116} {{Unsupervised Cross-Lingual
  Representation Learning at Scale}}.
\newblock \emph{arXiv preprint arXiv:1911.02116}.

\bibitem[{Conneau and Lample(2019)}]{conneau2019cross}
Alexis Conneau and Guillaume Lample. 2019.
\newblock \href
  {http://papers.nips.cc/paper/8928-cross-lingual-language-model-pretraining}
  {{Cross-Lingual Language Model Pretraining}}.
\newblock In \emph{Advances in Neural Information Processing Systems}, pages
  7059--7069.

\bibitem[{Cui et~al.(2019)Cui, Che, Liu, Qin, Yang, Wang, and
  Hu}]{cui2019pretraining}
Yiming Cui, Wanxiang Che, Ting Liu, Bing Qin, Ziqing Yang, Shijin Wang, and
  Guoping Hu. 2019.
\newblock \href {https://arxiv.org/abs/1906.08101} {{Pre-Training with Whole
  Word Masking for Chinese BERT}}.
\newblock \emph{arXiv preprint arXiv:1906.08101}.

\bibitem[{Delobelle et~al.(2020)Delobelle, Winters, and
  Berendt}]{delobelle2020robbert}
Pieter Delobelle, Thomas Winters, and Bettina Berendt. 2020.
\newblock \href {https://arxiv.org/abs/2001.06286} {{RobBERT: a Dutch
  RoBERTa-based Language Model}}.
\newblock \emph{arXiv preprint arXiv:2001.06286}.

\bibitem[{Devlin et~al.(2019)Devlin, Chang, Lee, and
  Toutanova}]{devlin2019bert}
Jacob Devlin, Ming-Wei Chang, Kenton Lee, and Kristina Toutanova. 2019.
\newblock \href {https://doi.org/10.18653/v1/N19-1423} {Bert: Pre-training of
  deep bidirectional transformers for language understanding}.
\newblock In \emph{Proceedings of the 2019 Conference of the North American
  Chapter of the Association for Computational Linguistics: Human Language
  Technologies, Volume 1 (Long and Short Papers)}, pages 4171--4186.

\bibitem[{Ginter et~al.(2017)Ginter, Haji{\v c}, Luotolahti, Straka, and
  Zeman}]{conll2017corpus}
Filip Ginter, Jan Haji{\v c}, Juhani Luotolahti, Milan Straka, and Daniel
  Zeman. 2017.
\newblock \href {http://hdl.handle.net/11234/1-1989} {{CoNLL 2017 Shared Task -
  Automatically Annotated Raw Texts and Word Embeddings}}.
\newblock {LINDAT}/{CLARIAH}-{CZ} digital library at the Institute of Formal
  and Applied Linguistics ({{\'U}FAL}), Faculty of Mathematics and Physics,
  Charles University.

\bibitem[{Kallas and Koppel(2018)}]{national2017corpus}
J.~Kallas and K.~Koppel. 2018.
\newblock \href
  {https://doi.org/https://doi.org/10.15155/3-00-0000-0000-0000-071E7L}
  {{Estonian National Corpus 2017}}.
\newblock Center of Estonian Language Resources.

\bibitem[{Kallas and Koppel(2019)}]{national2019corpus}
J.~Kallas and K.~Koppel. 2019.
\newblock \href
  {https://doi.org/https://doi.org/10.15155/3-00-0000-0000-0000-08565L}
  {{Estional National Corpus 2019}}.
\newblock Center of Estonian Language Resources.

\bibitem[{Kittask et~al.(2020)Kittask, Milintsevich, and
  Sirts}]{kittask2020evaluating}
Claudia Kittask, Kirill Milintsevich, and Kairit Sirts. 2020.
\newblock \href {https://doi.org/10.3233/FAIA200597} {{Evaluating multilingual
  BERT for Estonian}}.
\newblock In \emph{Volume 328: Human Language Technologies – The Baltic
  Perspective}, Frontiers in Artificial Intelligence and Applications, pages
  19--26.

\bibitem[{Kuratov and Arkhipov(2019)}]{kuratov2019adaptation}
Yuri Kuratov and Mikhail Arkhipov. 2019.
\newblock \href
  {http://www.dialog-21.ru/media/4606/kuratovyplusarkhipovm-025.pdf}
  {{Adaptation of Deep Bidirectional Multilingual Transformers for Russian
  Language}}.
\newblock \emph{Computational Linguistics and Intellectual Technologies},
  (18):333--339.

\bibitem[{Laur et~al.(2020)Laur, Orasmaa, Särg, and
  Tammo}]{laur-EtAl:2020:LREC}
Sven Laur, Siim Orasmaa, Dage Särg, and Paul Tammo. 2020.
\newblock \href {https://www.aclweb.org/anthology/2020.lrec-1.884} {{EstNLTK
  1.6: Remastered Estonian NLP Pipeline}}.
\newblock In \emph{Proceedings of The 12th Language Resources and Evaluation
  Conference}, pages 7154--7162.

\bibitem[{Le et~al.(2020)Le, Vial, Frej, Segonne, Coavoux, Lecouteux, Allauzen,
  Crabb{\'e}, Besacier, and Schwab}]{le2020flaubert}
Hang Le, Lo{\"\i}c Vial, Jibril Frej, Vincent Segonne, Maximin Coavoux,
  Benjamin Lecouteux, Alexandre Allauzen, Benoit Crabb{\'e}, Laurent Besacier,
  and Didier Schwab. 2020.
\newblock \href {https://doi.org/10.18653/v1/2020.acl-main.645} {{FlauBERT:
  Unsupervised Language Model Pre-training for French}}.
\newblock In \emph{Proceedings of The 12th Language Resources and Evaluation
  Conference}, pages 2479--2490.

\bibitem[{Martin et~al.(2020)Martin, Muller, Ortiz~Suárez, Dupont, Romary,
  de~la Clergerie, Seddah, and Sagot}]{Martin_2020}
Louis Martin, Benjamin Muller, Pedro~Javier Ortiz~Suárez, Yoann Dupont,
  Laurent Romary, {\'E}ric de~la Clergerie, Djamé Seddah, and Benoît Sagot.
  2020.
\newblock \href {https://doi.org/10.18653/v1/2020.acl-main.645} {{CamemBERT: a
  Tasty French Language Model}}.
\newblock \emph{Proceedings of the 58th Annual Meeting of the Association for
  Computational Linguistics}, pages 7203--7219.

\bibitem[{Nozza et~al.(2020)Nozza, Bianchi, and Hovy}]{nozza2020mask}
Debora Nozza, Federico Bianchi, and Dirk Hovy. 2020.
\newblock \href {https://arxiv.org/abs/2003.02912} {{What the [MASK]? Making
  Sense of Language-Specific BERT Models}}.
\newblock \emph{arXiv preprint arXiv:2003.02912}.

\bibitem[{Pajupuu et~al.(2016)Pajupuu, Altrov, and
  Pajupuu}]{Pajupuu2016IDENTIFYINGPI}
Hille Pajupuu, Rene Altrov, and Jaan Pajupuu. 2016.
\newblock {Identifying Polarity in Different Text Types}.
\newblock \emph{Folklore}, 64.

\bibitem[{Peters et~al.(2018)Peters, Neumann, Iyyer, Gardner, Clark, Lee, and
  Zettlemoyer}]{peters2018deep}
Matthew Peters, Mark Neumann, Mohit Iyyer, Matt Gardner, Christopher Clark,
  Kenton Lee, and Luke Zettlemoyer. 2018.
\newblock \href {https://doi.org/10.18653/v1/N18-1202} {{Deep Contextualized
  Word Representations}}.
\newblock In \emph{Proceedings of the 2018 Conference of the North American
  Chapter of the Association for Computational Linguistics: Human Language
  Technologies, Volume 1 (Long Papers)}, pages 2227--2237.

\bibitem[{Pyysalo et~al.(2020)Pyysalo, Kanerva, Virtanen, and
  Ginter}]{pyysalo2020wikibert}
Sampo Pyysalo, Jenna Kanerva, Antti Virtanen, and Filip Ginter. 2020.
\newblock \href {https://arxiv.org/abs/2006.01538} {Wikibert models: deep
  transfer learning for many languages}.
\newblock \emph{arXiv preprint arXiv:2006.01538}.

\bibitem[{Qi et~al.(2020)Qi, Zhang, Zhang, Bolton, and
  Manning}]{qi-etal-2020-stanza}
Peng Qi, Yuhao Zhang, Yuhui Zhang, Jason Bolton, and Christopher~D. Manning.
  2020.
\newblock \href {https://doi.org/10.18653/v1/2020.acl-demos.14} {{S}tanza: A
  python natural language processing toolkit for many human languages}.
\newblock In \emph{Proceedings of the 58th Annual Meeting of the Association
  for Computational Linguistics: System Demonstrations}, pages 101--108.

\bibitem[{Sanh et~al.(2019)Sanh, Debut, Chaumond, and
  Wolf}]{sanh2019distilbert}
Victor Sanh, Lysandre Debut, Julien Chaumond, and Thomas Wolf. 2019.
\newblock \href {https://arxiv.org/abs/1910.01108} {{DistilBERT, a distilled
  version of BERT: smaller, faster, cheaper and lighter}}.
\newblock \emph{arXiv preprint arXiv:1910.01108}.

\bibitem[{Tkachenko et~al.(2013)Tkachenko, Petmanson, and
  Laur}]{tkachenko2013named}
Alexander Tkachenko, Timo Petmanson, and Sven Laur. 2013.
\newblock \href {https://www.aclweb.org/anthology/W13-2412} {{Named Entity
  Recognition in Estonian}}.
\newblock In \emph{Proceedings of the 4th Biennial International Workshop on
  Balto-Slavic Natural Language Processing}, pages 78--83.

\bibitem[{{University of Tartu}(2018)}]{HPC}
{University of Tartu}. 2018.
\newblock \href {https://doi.org/10.23673/PH6N-0144} {{UT Rocket}}.
\newblock {share.neic.no}.

\bibitem[{Virtanen et~al.(2019)Virtanen, Kanerva, Ilo, Luoma, Luotolahti,
  Salakoski, Ginter, and Pyysalo}]{virtanen2019multilingual}
Antti Virtanen, Jenna Kanerva, Rami Ilo, Jouni Luoma, Juhani Luotolahti, Tapio
  Salakoski, Filip Ginter, and Sampo Pyysalo. 2019.
\newblock \href {https://arxiv.org/abs/1912.07076} {{Multilingual is not
  enough: BERT for Finnish}}.
\newblock \emph{arXiv preprint arXiv:1912.07076}.

\bibitem[{de~Vries et~al.(2019)de~Vries, van Cranenburgh, Bisazza, Caselli, van
  Noord, and Nissim}]{vries2019bertje}
Wietse de~Vries, Andreas van Cranenburgh, Arianna Bisazza, Tommaso Caselli,
  Gertjan van Noord, and Malvina Nissim. 2019.
\newblock \href {https://arxiv.org/abs/1912.09582} {{BERTje: A Dutch BERT
  Model}}.
\newblock \emph{arXiv preprint arXiv:1912.09582}.

\end{thebibliography}

\end{document}